\pdfoutput=1

\documentclass[11pt]{article}

\usepackage[review]{acl}

\usepackage{times}
\usepackage{latexsym}
\usepackage[review]{acl}
\usepackage{times}
\usepackage{latexsym}
\usepackage{tipa}
\usepackage{amsmath}
\usepackage{amsfonts} 
\usepackage{booktabs}
\usepackage{multirow}
\usepackage{adjustbox}
\usepackage{longtable}
\usepackage{graphicx}
\usepackage{float} 
\usepackage[T1]{fontenc}

\usepackage[utf8]{inputenc}

\usepackage{microtype}

\usepackage{inconsolata}

\usepackage{adjustbox}
\usepackage{afterpage}

\title{Multiple Sources are Better Than One: Incorporating External Knowledge in Low-Resource Glossing}

\author{
Changbing Yang, Garrett Nicolai, Miikka Silfverberg \\
University of British Columbia \\
{\tt cyang33@@mail.ubc.ca \quad garrett.nicolai@ubc.ca \quad miikka.silfverberg@ubc.ca}
}

\begin{document}
\nolinenumbers
\maketitle
\begin{abstract}
    
In this paper, we address the data scarcity problem in automatic data-driven glossing for low-resource languages by coordinating multiple sources of linguistic expertise. We supplement models with translations at both the token and sentence level as well as leverage the extensive linguistic capability of modern LLMs. Our enhancements lead to an average absolute improvement of 5\%-points in word-level accuracy over the previous state of the art on a typologically diverse dataset 
spanning six low-resource languages. The improvements are particularly noticeable for the lowest-resourced language Gitksan, where we achieve a 10\%-point improvement. Furthermore, in a simulated ultra-low resource setting for the same six languages, training on fewer than 100 glossed sentences, we establish an average 10\%-point improvement in word-level accuracy 
over the previous state-of-the-art system.
\end{abstract}

\section{Introduction}
The extinction rate of languages is alarmingly high, with an estimated 90\% of the world's languages at risk of disappearing within the next century~\cite{krauss1992world}. As speech communities dwindle, linguists are urgently prioritizing the documentation of these languages. This is a multi-step process involving: 1. phonetic and orthographic transcription, 2. translation into a so-called \textit{matrix language} like English or Spanish, which provides a common frame of reference for all annotations, 3. morpheme segmentation, and 4. grammatical annotation \cite{crowley2007field}. The end-result is represented as Interlinear Glossed Text (IGT) 
like the Gitksan example below 
(see Appendix \ref{sec:appendix} for additional details):
\vskip.15cm
\begin{adjustbox}{width=\columnwidth}
    \begin{tabular}{ll}
        \textbf{Orthography:} & Ii hahla'lsdi'y goohl IBM \\
        \textbf{Segmentation:} & ii hahla'lst-'y goo-hl IBM \\
        \textbf{Gloss:} & CCNJ work-1SG.II LOC-CN IBM \\
        \textbf{Translation:} & And I worked for IBM.\\
    \end{tabular}
\end{adjustbox}\\

The traditional manual approach to language documentation, while thorough, is notably labor-intensive. This has spurred the development of automated tools leveraging machine learning for tasks such as word segmentation and glossing. For example, \newcite{moeller-hulden-2018-automatic} train neural models for automatic glossing of Lezgi, a Nakh-Daghestanian language. Their models deliver reasonable performance when trained on a small training set of 3,000 glossed tokens of Lezgi text. However, neural models are data-hungry and the small training set prevents the models from reaching their full potential. The most straightforward way to improve model performance would be to manually gloss more training data. However, as stated above, manual glossing is a very time-consuming process. Therefore, additional data sources should be considered.

\begin{figure} [h]
  \includegraphics[width=\columnwidth]{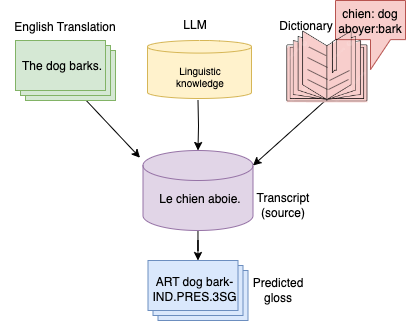}
  \caption{When glossing input such as the French sentence \textit{Le chien aboie}, our system utilizes multiple information sources: an English sentence-level translation, general linguistic knowledge provided by an LLM and dictionary definitions for the input tokens.}
  \label{fig:additem}
\end{figure}

Many recent glossing approaches \cite{girrbach-2023-tu, moeller-hulden-2018-automatic} exclusively train on glossed source language transcripts. However, we often have access to additional helpful knowledge sources. 
One option is to augment the data using translations of the training examples into the matrix language.\footnote{Frequently, the matrix language will be English but can also be another language like Spanish or Russian.} These provide an important source of lexical information because the gloss of nouns and verbs can often be found within the translation.\footnote{For the French sentence \textit{Le chien aboie}, the correct gloss of both \textit{chien} 'dog' and \textit{aboyer} 'bark' can be found in its English translation: \textit{The dog barks}.}
Because translation is a part of the language documentation process, these are often readily available 
and, thus, represent a quick and cost-effective way to provide an additional source of supervision.  Our system incorporates translations as an added information source. 

Unfortunately, the availability of translations for IGT data is necessarily limited simply because the quantity of IGT data itself is limited. 
As an additional source of lexical information, our system  incorporates external dictionaries which provide word-level translations of target language lexemes into the matrix language. 
This helps the system generalize to words missing from the training data. 

Recently, powerful pretrained models 
have emerged as a viable approach to strengthen and supplement the training signal for NLP tasks in low-resource settings \cite{ogueji2021small, bhattacharjee2021banglabert, hangya2022improving}. 
Advancements in large language models (LLMs) also present new opportunities for enhancing the language documentation process. Pretrained language models such as BERT \cite{devlin2018bert} and LLMs like GPT-4 \cite{achiam2023gpt}, trained on billions of tokens of text, encode extensive lexical and linguistic knowledge in the matrix language, and their incorporation has improved the benchmarks in many natural language tasks \cite{zhao2023survey, bommasani2021opportunities,zhou2023comprehensive}. We integrate LLMs into our glossing pipeline as a post-correction step through in-context learning. It is worth noting that our approach does not require fine-tuning and is, therefore, appropriate in low-resource settings where compute capacity is limited.

By leveraging three external sources of information (see Figure \ref{fig:additem}): utterance translations, external dictionaries and LLMs, our glossing pipeline achieves an average absolute improvement of 5\%-points over the previous state-of-the-art on datasets from the SIGMORPHON 2023 Shared Task on Interlinear Glossing \cite{ginn2023findings}. In particular, the incorporation of dictionaries leads to significant advancements for ultra-low resource languages such as Gitksan, resulting in a 10\%-points increase in word-level accuracy. 
Our key contributions are:

\noindent{\bf 1.} We enhance the training of glossing systems---in addition to plain glossed training examples, we introduce additional supervision in the form of input translations which are encoded using a pre-trained language model. 

\noindent{\bf 2.} 
We utilize external dictionaries which improve glossing performance, particularly for the lowest-resourced languages. 

\noindent {\bf 3.} We pioneer the use of LLM prompting and in-context learning techniques as a post-correction step 
in the glossing pipeline. To our knowledge, this is the first time LLMs have been applied to the automatic glossing task.
Our findings show that in-context prompting results in substantial improvements, especially when very limited training data is available. 

\section{Related Work}
\paragraph{Interlinear Glossing} 
Research into automatic glossing starts with rule-based analysis \cite{bender-etal-2014-learning, snoek-etal-2014-modeling} followed by data-driven neural models \cite{moeller-hulden-2018-automatic, girrbach-2023-tu, ginn2023taxonomic,zhao-etal-2020-automatic}. More recently, the integration of pre-trained multilingual models \cite{ginn2024glosslm, sheikh2024cmulab} has shown great potential to aid documentation projects. Our work is inspired by the success of these powerful models and aims to build upon their strengths.



\paragraph{Integrating Translation into the Glossing Task} 
We are not unique in incorporating translation information into a glossing system in the presence of small training datasets. The system presented by \newcite{okabe-yvon-2023-towards} is based on CRFs \cite{sutton2012introduction}, and also employs translations. However, in contrast to our approach, they heavily rely on source and target word alignments derived from an unsupervised alignment system \cite{jalili-sabet-etal-2020-simalign}. In low-resource settings, it is hard to learn an accurate alignment model.
\footnote{Moreover, \newcite{okabe-yvon-2023-towards} assume morphologically-segmented input, which considerably simplifies the glossing task. We instead address the much harder task of predicting glosses without segmentation information.} 

Pioneering studies by \newcite{zoph2016multi}, \newcite{anastasopoulos2018leveraging} and \newcite{zhao-etal-2020-automatic}, show that leveraging translations can enhance the performance of a neural glossing system. 
A notable limitation in all of these approaches is the scarcity of available English translations for training models. Therefore, only modest improvements in glossing accuracy are observed. Our work, in contrast, incorporates translation information through large pre-trained language models, which leads to greater improvements in glossing performance. 
This strategy has lately become increasingly popular in low-resource NLP and shows promise across various language processing tasks \cite{ogueji2021small, hangya2022improving}.

Similarly to our approach, \newcite{okabe-yvon-2023-towards} also take advantage of the BERT model in their study, but only utilize BERT representations for translation alignment. In contrast, we directly incorporate encoded translations into our glossing model. 
\newcite{he2023sigmorefun} also use pre-trained language models, namely, XLM-Roberta \cite{conneau2020unsupervised}, mT5 \cite{xue2021mt5} and ByT5 \cite{xue2022byt5}, as part of their glossing model. However, they do not incorporate IGT translation information.\footnote{Though \newcite{he2023sigmorefun} do use external dictionary information for post-correction of glosses.} Instead, they directly fine-tune the pre-trained models for glossing.

\paragraph{LLM Prompting}
In recent years, the application of LLMs for various NLP tasks has expanded significantly, demonstrating remarkable potential in few-shot and in-context learning. This approach leverages the inherent knowledge and adaptability of LLMs like GPT-4 \cite{achiam2023gpt} and LLaMA-3 \cite{touvron2023llama}, allowing them to perform tasks based on a few examples provided as context, without requiring further fine-tuning. \newcite{margatina-etal-2023-active} introduce a novel perspective by applying active learning (AL) principles to in-context learning with LLMs. Their study frames the selection of in-context examples as a pool-based AL problem conducted over a single iteration. Various AL algorithms, including uncertainty, diversity, and similarity-based sampling, is explored to identify the most informative examples for in-context learning. The findings consistently indicate that selecting examples semantically similar to the test instances significantly outperforms other methods, including random sampling and traditional uncertainty-based approaches .

Building on these insights, our proposed work aims to enhance the task of automatic glossing in low-resource settings by integrating LLM prompting and active learning principles. Our approach applies the strategies outlined by \cite{margatina-etal-2023-active} by focusing on similarity-based methods for selecting in-context examples. This ensures that the most relevant and informative examples are utilized, enhancing the model's ability to generate accurate glosses. Additionally, we explore the effectiveness of various active learning methods such as BERT-similarity, word overlapping, longest common subsequence, and random sampling, tailoring these approaches to the specific needs of the glossing task.

\label{experiment}

\begin{figure*}
    \centering
    \begin{minipage}{0.2\textwidth}
        \centering
        \includegraphics[width=0.8\textwidth]{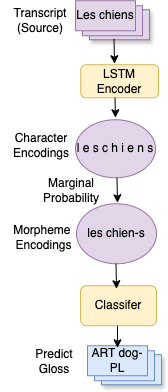} 
        \caption{Pipeline of \newcite{girrbach-2023-tu}'s model.}
     \label{fig:baselinepipeline}
    \end{minipage}\hfill
    \begin{minipage}{0.76\textwidth}
        \centering
        \includegraphics[width=0.9\textwidth]{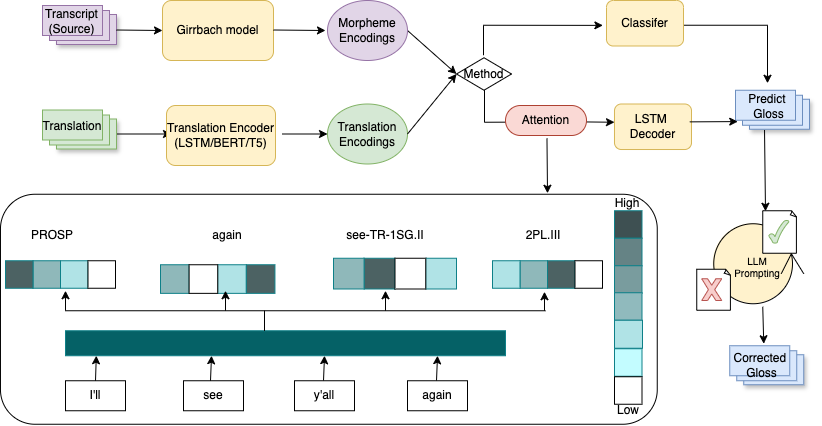} 
        \caption{Pipeline of the proposed work. The lower portion of the diagram demonstrates how attention weights inform the model when predicting the glossing targets.}
        \label{fig:pipeline2}
    \end{minipage}
\end{figure*}

\section{Data}
We conduct experiments on data from the 2023 SIGMORPHON shared task on interlinear glossing \cite{ginn2023findings}. 
The shared task provides two distinct tracks: an open track, where the input is morphologically segmented, and a closed track, where no segmentations are provided.
Our analysis focuses on data from the closed track. This setting is substantially more challenging because morphological segmentation now, effectively, becomes a part of the glossing task. 
The closed-track languages are Arapaho (arp), Gitksan (git), Lezgi (lez), Nat\"ugu (ntu), Tsez (ddo), and Uspanteko (usp).\footnote{We exclude one language Nyangbo, because its dataset lacks translations.} Data details are shown as in Table \ref{tab:data-detail-1}. 
With most languages, except Arapaho, comprising fewer than training 10,000 sentences, our datasets can be called low-resourced. 
For all languages, the data includes translations in a matrix language which is English, except from Uspanteko, where it is Spanish. 

\begin{table}
\begin{adjustbox}{width=\columnwidth,center}
\begin{tabular}{lllll}
\toprule
Language & Train(num) & Dev(num) & Test(num) & Matrix lang.  \\\midrule
Arapaho (arp) & 39,501 & 4,938 & 4,892 & (eng) \\
Gitksan (git) & 31 & 42 & 37 & (eng) \\
Lezgi (lez) & 701 & 88 & 87 & (eng) \\
Nat\"ugu (ntu) & 791 & 99 & 99 & (eng) \\
Tsez (ddo) & 3,558 & 445 & 445 & (eng) \\
Uspanteko (usp) & 9,774 & 232 & 633 & (spa) \\
\bottomrule
\end{tabular}
\end{adjustbox}
\caption{2023 Sigmorphon Shared Task Dataset Information \cite{ginn2023findings}}
\label{tab:data-detail-1}
\end{table}

\section{Baseline Model} 

Our glossing system is based upon a neural glossing model developed by \newcite{girrbach-2023-tu}. This is the winning system of the 2023 SIGMORPHON shared task on internlinear glossing. As shown in Figure \ref{fig:baselinepipeline}, the model accomplishes glossing of morphological segments through a three-stage process: input encoding, unsupervised morpheme segmentation, and morpheme classification. 

\paragraph{Input encoder} The model input consists of a character-sequence $s = s_1,\ ...,\ s_N$, representing a sentence. A bidirectional long short-term memory network (BiLSTM) encodes the input into a sequence of contextualized embeddings $\mathbf{h}_i$, one for every character in $s$. 

\paragraph{Morpheme Segmenter} Next, the model performs unsupervised morphological segmentation using the forward-backward algorithm \cite{kim2016structured}. 
In a first step, an MLP is used to  predict the number of morphemes $J_w$ for each word $w$ in input sentence $s$. 
For each character $s_i$, the model applies a linear layer with Sigmoid activation function to its character encoding $\mathbf{h}_i$ to get the probability $p_i^{\text{seg}}$ that indicates whether $s_i$ is the last character of the morpheme segment. Then the forward and backward scores ($\alpha$ and $\beta$, respectively) for each input position $i$ and target morpheme $j$ can be computed as follows:
\[
\alpha_{i,j} = \alpha_{i-1,j} \cdot (1 - p^{\text{seg}}_{i-1}) + \alpha_{i-1,j-1} \cdot p^{\text{seg}}_{i-1}
\]
\[
\beta_{i,j} = \beta_{i+1,j} \cdot (1 - p^{\text{seg}}_i) + \beta_{i+1,j+1} \cdot p^{\text{seg}}_i
\]
Finally, the marginal probability of a morpheme boundary at position $i$ relating to morpheme $j$ is given by:

\[
\xi_{i,j} = \frac{\alpha_{i,j} \cdot \beta_{i,j}}{\alpha_{N,J_w}}
\]
where $N$ is the sequence length, and $J_w$ is the number of morphemes in the word $w$.

\paragraph{Morpheme classifier} 
After segmentation, we get each morpheme encoding $\mathbf{e_j}$ through averaging its corresponding character encodings. An MLP is then used to predict the gloss for each morpheme based on its morpheme encoding. Model training optimizes the cross-entropy loss between the predicted and ground-truth gloss labels. 


\section{Our Methods}
\label{our-model}
Our glossing system enhances the baseline model by incorporating utterance translations (both at the sentence level and token level) and a character-based decoder.\footnote{Our code is publicly available: \url{https://github.com/changbingY/Auto_glossing_stem_translation}} Model and training details are provided in Appendix \ref{model_setting}. Additionally, we implement a gloss post-correction component using LLM-powered in-context learning. Figure \ref{fig:pipeline2} presents an overview of the system.

\subsection{Character-Based Gloss Decoder} Our first addition to the \newcite{girrbach-2023-tu} model is a character-based decoder. The baseline model is unable to predict glosses which were not observed in the training data, because it treats glossing as a morpheme classification task with a closed set of potential gloss labels. This deficiency is particularly harmful when predicting glosses for lexical morphemes (i.e. word stems) which represent a much larger inventory than grammatical morphemes (i.e. inflectional and derivational affixes). A character-based decoder can enhance the model's capability to use words from a translation of the input example. 
Following \newcite{kann2016med}, we implemented a LSTM decoder. However, we adapt it to function at the character level for lexical morphemes and at the morpheme level for grammatical morphemes. \footnote{For instance, if the word gloss is "dog-FOC", the decoder will generate it as "d-o-g-FOC".}

\subsection{Translation Encoder}
We then extend the model of \newcite{girrbach-2023-tu} by incorporating matrix-language translations. We encode the English or Spanish (in the case of Uspanteko) translations in the shared task datasets using a deep encoder. We experiment with three different encoders: a character-based BiLSTM \cite{hochreiter1997long} and pre-trained transformers BERT-base \cite{kenton2019bert} and T5-large \cite{raffel2020exploring}.\footnote{See Appendix \ref{model_setting} for details concerning the encoders.} To represent translations, we then either use the final hidden state from the translation encoder, or attend over the translation hidden states. 


When attending over the hidden states, we apply Bahdanau attention \cite{bahdanau2014neural} scoring the association between each encoder hidden states and the previous decoder state $\mathbf{d}_{i-1}$. We separately attend to the encoded morpheme representations $\mathbf{e_j}$ in the input example (morphemes are discovered by our baseline model in an unsupervised manner as explained above) and the encoded subword-tokens $\mathbf{t_k}$ in the translation. 
This gives us a morpheme representation $\mathbf{e}_i = \sum_{j=1}^J w^e_j \mathbf{e_j}$ and a translation representation $\mathbf{t}_i = \sum_{k=1}^K w^t_k \mathbf{t_k}$ at time-step $i$. We then use the concatenated representation $[\mathbf{e}_i; \mathbf{t}_i]$
to compute the next gloss decoder state $\mathbf{d}_{i}$.    



\subsection{Post-correction through in-context learning}

Preliminary experiments revealed that the glossing system sometimes generates typos and non-sensical glosses such as \textit{stoply} instead of \textit{story}. To mitigate this issue, we introduce a post-correction step leveraging LLM prompting. We enhance the accuracy and reliability of glosses through an in-context learning approach. 


For each language, we generate conservative silver glosses (requiring correction) using a BERT-based model with attention (BERT+attn+chr) to prevent excessive corrections, as the baseline model \cite{girrbach-2023-tu} already provides a reasonably accurate starting point. We use one-quarter of the training data to produce silver glosses for the remaining training data, fine-tuning the model on the original development split. To reduce noise, we apply an edit distance constraint, retaining examples where the gloss edit distance from the gold gloss is limited to 4-8 characters.\footnote{The character number is determined by half the length of the word glosses, depending on the language.} The initial one-quarter of data is then reintroduced into the training set, ensuring completeness and accuracy, as these glosses match the original training data.

\begin{figure*}
\includegraphics[width=\textwidth]{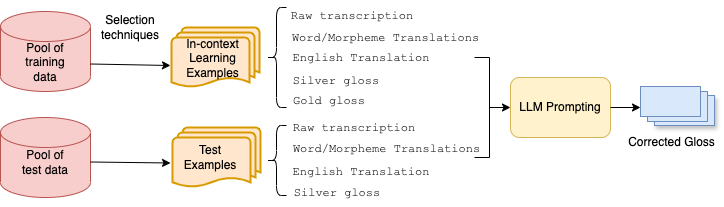}
\caption{The procedure of selecting in-context learning examples to generate components for LLM prompting.}
\label{prompt-select}
\end{figure*}

Here we prepare a prompt which asks the LLM to correct the lexical morphemes in a glossed input sentence. A prompt is generated by selecting two training examples as in-context learning examples for each test example. Each in-context learning example includes the source language transcript, morpheme/word translations based on the training data, the English translation of the sentence, the silver gloss, and the gold gloss. The test example is structured similarly but omits the gold gloss, prompting the language model to generate the corrected gloss. The prompting pipeline is illustrated in Figure \ref{prompt-select}. When using an external dictionary, we additionally provide word translations in the prompt. Following the in-context paradigm, we do not perform any further training or fine-tuning of the LLM. The template used for the prompting is detailed in Appendix \ref{template}.
We experiment with two models in this scenario: GPT-4 \cite{achiam2023gpt} and LLaMA-3 \cite{touvron2023llama}.

\textbf{In-context Learning Examples Selection Techniques}
In our experiment, we compare three techniques to optimize the selection of in-context learning examples. We evaluate these techniques against random selection.
\textbf{BERT Similarity (BERT-Sim)}
We first embed the translated test sentence from the IGT using BERT (we use multilingual BERT for Uspanteko).  We then find the two training sentences with the lowest embedded cosine distance from the test case, and use them as our in-context examples.
\textbf{Overlapping Words (Overlap)}
We calculate the number of overlapping words between source sentences in the test and training datasets.  In-context examples are selected to maximize the number of overlapping words between the test case and the training sentences.
\textbf{Longest Common Substrings (LCS)}
We select in-context examples from the training sentences that maximize the LCS with the test case.

\section{Experiments and Results}
In all experiments, we evaluate based on word-level glossing accuracy.
\subsection{Translation Enriched Model Results}
\begin{table*}[]
\begin{adjustbox}{width=\textwidth,center}
\begin{tabular}{lllllll|lllllll}
\toprule
Model setting        & arp  & lez  & ntu  & ddo & usp & ave&  arp-low  & git-low & lez-low  & ntu-low   & ddo-low  & usp-low &ave      \\\midrule
\newcite{girrbach-2023-tu} & 78.79& 78.78  & 81.04  & 80.96 & 73.39 & 78.59 &19.12 & 21.09 & 48.84  & 51.08 & 36.12 & 17.32 & 32.26 \\\bottomrule
LSTM & 77.04& 81.42 & 83.55 & 84.99 & 73.01 & 80.00&18.67 & 20.71 &54.29  & 59.56 & \textbf{44.5}  & 32.92 & 38.44\\
LSTM+attn  & 79.31 & 76.19 & 83.01 & 85.12 & 76.24 &79.97 &24.38 &  18.49 &  55.75 & 58.48 & 42.37 & 29.52 & 38.17\\
BERT+attn  & 78.98 & 81.87 & 84.57 & 85.84 & 77.63 &81.78 & 27.33 &  20.31  & 55.86  & 60.13  & 41.85 & 33.04 & 39.75\\
BERT+attn+chr  & 80.79 & 82.19 & \textbf{85.41} & 84.13 & \textbf{79.34}   & 82.37&\textbf{28.82} & \textbf{28.11} & 56.99 & 62.73  & 39.72 & \textbf{35.84} & \textbf{42.04}   \\
T5+attn+chr & \textbf{81.11}  & \textbf{82.37} & 84.68 & \textbf{85.91} & 78.72 & \textbf{82.56} & 27.31  & 24.23 & \textbf{57.33} & \textbf{62.82} & 39.97 & 33.59& 40.88 \\\bottomrule
\end{tabular}
\end{adjustbox}
\caption{Word-level accuracy of languages in the 2023 Sigmorphon Shared Task \cite{ginn2023findings} (left) and ultra low-resource settings (right). Model specifics are elaborated in Section \ref{our-model}. }
\label{app:gloss_stats-encodeco}
\end{table*}

Table \ref{app:gloss_stats-encodeco} shows the glossing accuracy across different model settings and languages.\footnote{We additionally present edit distance in Appendix \ref{app:edit-distance}.} We report performance separately for original shared task datasets and our simulated ultra low-resource datasets spanning 100 training sentences. We group the Gitksan shared task dataset in the ultra low-resource category because it only has 30 training examples.\footnote{Apart from the baseline, all systems apply majority voting from 10 independently trained models. Its impact is discussed in  Appendix \ref{majorityvoting}.} 

\paragraph{Shared Task Data} 
When only integrating translations through the final state of a bidirectional LSTM, we observe an improvement in average glossing accuracy, but performance is reduced for two languages (Arapaho and Uspanteko). 

Augmenting translations via an attentional mechanism (LSTM+attn) does not confer consistent improvements. 
In contrast, translation information incorporated via a pre-trained model (BERT+attn) renders consistent improvements in glossing accuracy across all languages and we see notable gains in average glossing accuracy over the baseline.  
Incorporating a character-based decoder leads to further improvements in average glossing accuracy and for all individual languages. The T5 model (T5+attn+chr) attains the highest average performance: 82.56\%, which represents a 3.97\%-points improvement over the baseline. It also  delivers the highest performance for three out of our five test languages (Arapaho, Lezgi and Tsez), while the BERT-based model with attention (BERT+attn+chr) delivers the best performance for the remaining two (Nat\"ugu and Uspanteko). 
Among all languages, we see improvements over the baseline model ranging from 2.32\%-points to 5.95\%-points.\footnote{We visualize the attention patterns over
the English translation representations. The visualizations are shown in Appendix \ref{attn-distribution-graphs-appendix}}

\paragraph{Ultra Low-Resource Data} 
In order to investigate the performance of our model in ultra low-resource settings, we additionally form smaller training sets by sampling 100 sentences from the original shared task training data. We use the original shared task development and test sets for validation and testing, respectively. 

Translations integrated through the final state of a randomly initialized bidirectional LSTM  (LSTM and LSTM+attn),
lead to an average 6\%-points improvement in accuracy over the baseline. We achieve particularly impressive gains for
Uspanteko, surpassing the baseline accuracy by over 15\%-points. Incorporating pre-trained models (BERT+attn) 
exhibits a slight increase in accuracy for certain languages. 
However, when we incorporate both pre-trained models and the character-based decoder (BERT+attn+chr and T5+attn+chr), we see larger gains in accuracy
across the board. Here, BERT achieves the highest average accuracy of 42.04\%, which represents a 9.78\%-points improvement over the baseline. It achieves the highest performance for three languages (Arapaho, Gitksan and Uspanteko), while T5 delivers the best performance for two of the languages (Lezgi and Nat\"ugu). The plain LSTM model attains the best performance for Tsez.

\subsection{Prompting Model Results}

\begin{table}

\begin{adjustbox}{width=\columnwidth,center}
\begin{tabular}{lllllll}
\toprule
Model setting        & arp  & lez  & ntu  & ddo & usp & git       \\\midrule
T5/BERT+attn+chr& 81.11  & 82.37 & 85.41 & \textbf{85.91} & \textbf{79.34} & 28.11  \\
+GPT4-random & 81.12 & 83.52& 85.79& 84.76 & 70.62& 28.58 \\
+GPT4-BERT-Sim & 81.17 & \textbf{84.70}& 86.07 & 85.32& 72.44&29.02 \\
+GPT4-Overlap & \textbf{81.57} & 84.47&  86.11&  85.53  &  73.64 & 29.14 \\
+GPT4-LCS & 81.25 &83.86 &\textbf{86.38} & 84.98 & 72.78& 28.77 \\
+LLaMA3-Overlap & 81.23 & 83.01& 86.09 & 83.77 &70.99 &  \textbf{30.11}\\\bottomrule

\end{tabular}
\end{adjustbox}
\caption{Word-level accuracy of all languages. We incorporate prompts using different selection techniques for in-context examples, which add into the information enriched models (T5/BERT+attn+chr). }
\label{app:prompt-word-combine-tech}

\begin{adjustbox}{width=0.5\textwidth,center}
\begin{tabular}{llllllll}
\toprule
Model setting        & arp  & lez  & ntu  & ddo & usp & git &ave      \\\midrule
\newcite{girrbach-2023-tu} & 78.79& 78.78  & 81.04  & 80.96 & 73.39 &21.09 &69.01  \\
T5/BERT+attn+chr& 81.11  & 82.37 & 85.41 & \textbf{85.91} & \textbf{79.34} & 28.11 & \textbf{73.88} \\
T5/BERT+attn+chr+Prmpt & \textbf{81.57} &\textbf{84.70} & \textbf{86.38} & 85.53 & 73.64 & \textbf{30.11} &73.66  \\\bottomrule
\end{tabular}
\end{adjustbox}
\caption{Word-level accuracy of all languages. We compare the performance of models that incorporate prompts from our optimal in-context example selection techniques with other models. }
\label{app:prompt-word1}
\end{table}

The prompting experiments aim to further improve the output of the T5/BERT+attn+chr model by post-correcting its glossed output using an LLM. We only allow the LLM to change the gloss of lexical morphemes because preliminary experiments demonstrated that post-processing tends to worsen performance on grammatical morphemes. The word-level accuracy shown in Table \ref{app:prompt-word-combine-tech} highlights the performance of various training data selection techniques across multiple languages.\footnote{We additionally present lexical morpheme accuracy in Appendix \ref{lex-technique}.} We further select the best setting to compare with the baseline model and translation enriched models. The comparison demonstrates that using in-context learning continues to boost glossing accuracy. This approach delivers further improvements for Arapaho, Lezgi, Nat\"ugu, and Gitksan. It presents the highest accuracy for Lezgi , showing a 2.33\%-points increase over the highest-performing translation enriched model T5/BERT+attn+chr.

When applying GPT-4 for post-correction, the Overlapping Words selection technique emerges as the most effective, achieving the highest accuracy for Arapaho at 81.57\% and maintaining strong performance across other languages. The BERT similarity and LCS techniques also provide substantial improvements over random selection, with notable improvements for Lezgi at 84.70\% and Nat\"ugu at 86.38\% accuracy, respectively. Additionally, the LLaMA-3 model using the Overlapping Words method shows competitive results, particularly excelling in the low-resource language Gitksan at 30.11\%, indicating its potential utility in such challenging settings.

We further examine predictions from the prompting model. One such example in Lezgi includes a sentence whose translation is "\textit{She was lonely}". The pre-corrected gloss from our encoder-decoder model (T5/BERT+attn+chr) contains incorrect lexical morpheme glosses, including ``pie'' and ``he''. It is evident that the prompting model successfully changed these lexical morphemes according to the words in the translation line of the IGT\footnote{We observe that the prompting results can contain synonyms. To gain a better understanding of our model's performance, we use BERT score as an alternative evaluation metric to evaluate the lexical morphemes. Results are shown in Appendix \ref{bert-score}.}. Results are as shown below:

\begin{adjustbox}{width=\columnwidth, center}

    \begin{tabular}{ll}
    & \\
        \textbf{Silver Gloss:}& {\color{red}pie} old.woman was {\color{red}he} \\
         \textbf{Prompt Gloss:} & {\color{red}alone} old.woman was {\color{red}still.be} \\
         \textbf{Gold Gloss:}& {\color{red}alone} old.woman was {\color{red}still.be,.remain} \\
    & \\ 
    \end{tabular}

\end{adjustbox}

Interestingly, both the GPT-4 and LLaMA-3 in-context learning setups perform worse when the translations are in Spanish than in English, as evidenced by the accuracy drop in Uspanteko. The reasons behind this require further investigation.

\subsection{External Dictionaries} We also assess the impact of introducing additional word translations into the in-context prompts to enhance accuracy.
We expand the word translations in the prompt using word translations from an external dictionary for Arapaho, Lezgi, and Gitksan. The source and detailed information about the dictionaries are shown in Appendix \ref{dictionary-source}. 
The word-level results, as presented in Table \ref{app:prompt-word-dict1}
illustrate that the integration of out-of-domain dictionary resources is highly beneficial, especially for languages with limited training data like Gitksan. Dictionary translations consistently boost the performance of our best models, enhancing benefits obtained solely through prompting. The dictionary-supplemented models achieve the best results in all three languages, with an overall average accuracy of 66.08\%, surpassing the baseline model by 6.53\%-points and the plain prompting model by 0.62\%-points.

\begin{table}[]

\begin{adjustbox}{width=0.5\textwidth,center}
\begin{tabular}{llllllll}
\toprule
Model setting        & arp  & lez  & git &ave    \\\midrule
\newcite{girrbach-2023-tu} & 78.79& 78.78  &21.09 & 59.55 \\
T5/BERT+attn+chr& 81.11  & 82.37  & 28.11 & 63.86  \\
T5/BERT+attn+chr+Prmpt & 81.57&84.70 &  30.11 & 65.46
\\
T5/BERT+attn+chr+Prmpt+Dict & \textbf{81.61} &\textbf{85.30} &  \textbf{31.32} &\textbf{66.08}\\\bottomrule
\end{tabular}
\end{adjustbox}
\caption{
Word-level accuracy of all languages. We compare the model performance among the accumulated effort of incorporating external dictionaries with other models.}
\label{app:prompt-word-dict1}
\end{table}

\subsection{Learning Curves}
\begin{figure}[H]
  \includegraphics[width=\columnwidth]{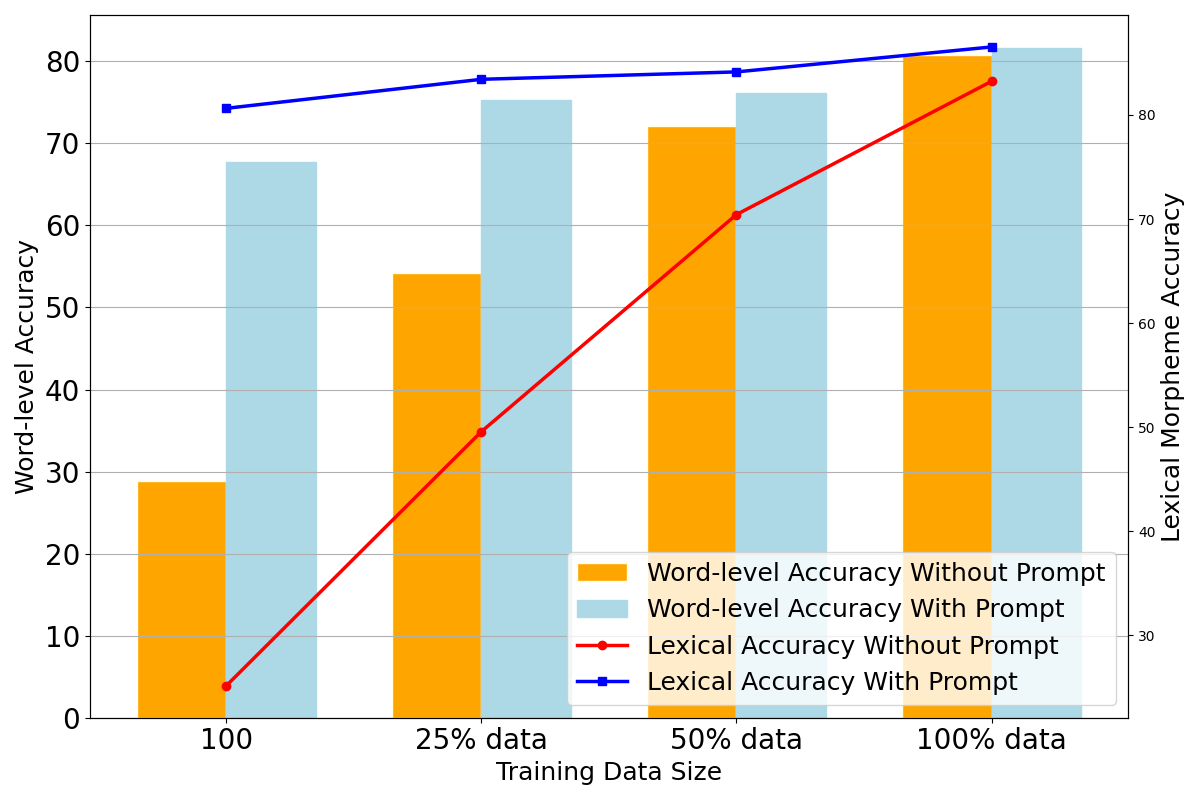}
  \caption{Lexical morpheme and word-level accuracy on Arapaho. We incorporate prompting with the encoder-decoder model which is enriched with translation.}
  \label{fig:train-size}
\end{figure}

The learning curves in Figure \ref{fig:train-size} illustrate the impact of prompting on model performance when using varying amounts of IGT training data. This comparison includes models with and without prompting, focusing on both word-level and lexical morpheme accuracy. We focuse on the Arapaho language, which has the largest number of manually glossed training examples: 39,501 training sentences, in total.

The bar chart represents the word-level accuracy for models trained with varying amounts of data (100 sentences, 25\% data, 50\% data, and 100\% data). The results clearly demonstrate that in-context post-correction greatly improves glossing accuracy. In ultra-low data conditions, the post-corrected model is more than twice as accurate as the uncorrected model.  As the amount of training data increases, the benefits gained through prompting diminish.

The line chart maps the accuracy of lexical morphemes prior- and post-correction. Similarly to the word-level accuracy, the accuracy of lexical morphemes benefits greatly from in-context post-correction. The most significant improvements are again observed when training data is restricted. With only 100 training sentences, the post-corrective model achieves a lexical morpheme accuracy that is nearly as high as that obtained using the full dataset.

\section{Conclusions}
This paper offers a promising and efficient solution by introducing multiple resources to aid in the glossing task, particularly in linguistically diverse and data-sparse environments. The current study demonstrates the effectiveness of incorporating translation information at both the token and sentence level, alongside LLM prompting in automatic glossing for low-resource languages. The proposed system, based on a modified version of Girrbach's model \cite{girrbach-2023-tu}, shows significant performance enhancements, particularly in low-resource settings. By leveraging translation data and integrating a character-based decoder, our approach provides a robust solution for unobserved lexical morphemes (stems).

This research pioneers the application of LLM prompting to the glossing task. By employing various in-context example selection strategies and adding extra dictionary words as a resource, we have shown that LLM prompting can substantially refine lexical morpheme glosses, leading to higher word-level accuracy. This approach is also particularly beneficial in scenarios with limited training data, as it maximizes the potential of minimal data resources.

In all, the integration of translation information, additional dictionary resources, along with LLM prompting, sets a new benchmark in automatic glossing.

\section{Limitations}
The limitations of our study primarily pertain to the extent of our experimentation and the models we have chosen. Firstly, our investigation relies solely on an LSTM decoder. This decision was influenced by time constraints, which limited our ability to explore more complex decoders. Additionally, our experimentation is confined to the T5-large model. While this model has shown promising results in our study, we acknowledge the existence of other large language models in the field of natural language processing. Although we did explore other large language models such as LLaMA-2 \cite{touvron2023llama}, our preliminary experiments yielded unsatisfactory results compared to T5. Consequently, we made the decision not to include LLaMA-2 in our paper due to its inferior performance. These limitations underscore the need for future research to explore a wider range of decoding architectures and incorporate various large language models to enhance our understanding of the subject matter. However, using large language models requires significant computational resources, which can have an environmental impact due to increased energy consumption. 

\bibliography{anthology}


\appendix
\section{IGT Information}
\label{sec:appendix}
In the IGT data, the second line includes segmentations with morphemes normalized to a canonical orthographic form. The third line has an abbreviated gloss for each segmented morpheme. Lexical morphemes typically correspond to the stems of words. The morpheme glosses usually have two categories: Lexical and Grammatical morphemes. For example, in glossing labels such as work-1SG.II, ``work" would be considered a Lexical morpheme, representing the core semantic unit. On the other hand, Grammatical morphemes like `1SG.II" are often denoted by uppercase glosses and generally signify grammatical functions, such as tense, aspect, or case, rather than specific lexical content.



\section{Model Settings}
\label{model_setting}
Our experimental framework and hyperparameters draw inspiration from Girrbach's methodology, with a focus on organizing and optimizing the technical setup. For model optimization, we employ the AdamW optimizer \cite{loshchilov2017decoupled}, excluding weight decay, and set the learning rate at 0.001. Except for this specific adjustment, we maintain PyTorch's default settings for all other parameters.

Our configuration is structured to allow a range of experiments, varying from 1 to 2 LSTM layers, with hidden sizes spanning from 64 to 512, and dropout rates fluctuating between 0.0 and 0.5. The scheduler $\gamma$ is adjusted within a range of 0.9 to 1.0, and batch sizes are diversified, ranging from 2 to 64. This versatile approach is designed to thoroughly evaluate the model's performance across a spectrum of hyperparameter configurations.

Departing from the original model which was trained for 25 epochs, our approach extends the training duration to 300 epochs when using large pretrained models. In cases where the BERT model is utilized, we sometime apply a 0.5 dropout rate during the BERT training phase. We exclusively employ the multilingual BERT model for Uspanteko, while we utilize the standard BERT model for all other languages. This comprehensive and meticulously organized setup is aimed at enhancing the effectiveness and efficiency of our model training process. 

To prevent coincidences, for each proposed model configuration, we train the model for 10 iterations, and the final prediction is determined through majority voting.

\section{Edit Distance}
\label{app:edit-distance}
Results are shown in Table \ref{app:gloss_stats_edit_distance}.
\begin{table*}[]
\begin{adjustbox}{width=\textwidth,center}
\begin{tabular}{lllllll|lllll}
\toprule
Model setting        & ara & git(-low)  & lez  & ntu  & ddo & usp &  ara-low   & lez-low  & ntu-low   & ddo-low  & usp-low       \\\midrule
\newcite{girrbach-2023-tu} & -&- &- &- &- &- &6.59   &3.64  & 4.78 & 4.92 & 3.79\\
LSTM & 1.52 & 5.65  &1.22  & \textbf{1.17 } & 0.72 & 0.88 &6.50& 3.28 & 4.12 & \textbf{3.93}  &  2.84 \\
LSTM+attn & \textbf{1.31}& 6.27  &  1.62&  1.34& 0.72 & 0.86 & 6.04   & 3.26 & 3.81 & 4.25 &3.21  \\
BERT+attn & 1.39 & 5.57  & 1.24 &  1.23&  0.69&  \textbf{0.70} & 5.97 &3.20  & 3.81  &  4.1& 2.88\\
BERT+attn+chr & 1.50& \textbf{5.30} & 1.20 & 1.25 & 0.53 & 0.81 &\textbf{5.54} & 3.04  & \textbf{3.55}   & 4.27 & 2.78  \\
T5+attn+chr &1.40 & 5.51 & \textbf{1.18} & 1.27 & \textbf{0.52}  & 0.78 &5.62 & \textbf{3.00}  & \textbf{3.55}   & 4.36  & \textbf{2.74}  \\\bottomrule
\end{tabular}
\end{adjustbox}
\caption{Word-level edit distance of languages in the 2023 Sigmorphon Shared Task \cite{ginn2023findings} (left) and low-resource settings (right), with `arp' representing Arapaho, `git' for Gitksan, `lez' for Lezgi, `ntu' for Nat\"ugu, `ddo' for Tsez, and `usp' for Uspanteko. Model specifics are elaborated in Section \ref{experiment}. }
\label{app:gloss_stats_edit_distance}
\end{table*}

\section{Influence of Majority Voting}
\label{majorityvoting}
Average accuracy across 10 models and results utilized majority voting are shown in Table \ref{app:majority}. Improvements in performance can be achieved even without resorting to voting, particularly accentuated in ultra low-resource datasets as opposed to the Shared Task datasets.
\begin{table*}[]
\begin{adjustbox}{width=\textwidth,center}
\begin{tabular}{lllllll|lllllll}
\toprule
Model setting        & arp  & lez  & ntu  & ddo & usp & ave&  arp-low  & git-low & lez-low  & ntu-low   & ddo-low  & usp-low &ave      \\\midrule
\newcite{girrbach-2023-tu} & 78.79& 78.78  & 81.04  & 80.96 & 73.39 & 78.59 &19.12 & 21.09 & 48.84  & 51.08 & 36.12 & 17.32 & 32.26 \\\bottomrule

BERT/T5+attn+chr-average & 79.32 & 79.49 & 80.76 & 81.00 & 74.92  & 79.10 &25.43 & 23.95 & 54.28& 57.18 & 32.41 & 28.77 &  37.00 \\
BERT/T5+attn+chr-majority & \textbf{81.11}  & \textbf{82.37} & \textbf{85.41} & \textbf{85.91} & \textbf{79.34} & \textbf{82.83} &\textbf{28.82}  &  \textbf{28.11} & \textbf{57.33} & \textbf{62.82} & \textbf{39.97} &\textbf{35.84} &  \textbf{42.14} \\\bottomrule
\end{tabular}
\end{adjustbox}
\caption{Word-level accuracy of languages in the 2023 Sigmorphon Shared Task \cite{ginn2023findings} and low-resource settings. We compute the average across 10 models and also utilized majority voting accuracy results. Language abbreviations were used, with `arp' representing Arapaho, `git' for Gitksan, 
 `lez' for Lezgi, `ntu' for Nat\"ugu, `ddo' for Tsez, and `usp' for Uspanteko. Model specifics are elaborated in Section \ref{experiment}. }
\label{app:majority}
\end{table*}

\section{Attention Distribution}

\label{attn-distribution-graphs-appendix}
\begin{figure}[h]
  \begin{minipage}{\columnwidth}
    \centering
    \includegraphics[width=\textwidth]{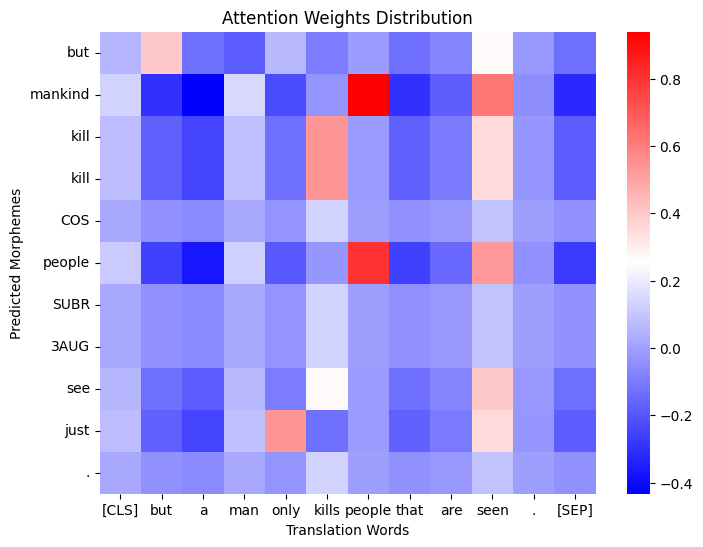}
  
  \end{minipage}
  \caption{Difference from mean attention weights of glossed output tokens (y-axis) with respect to encoded translation tokens (x-axis) for a Nat\"ugu example (attention weights are derived from the model BERT+attn+chr).}

  \label{fig:attdis-natugu-bert1}

\end{figure}

\begin{figure*}[htbp]
    \centering
    \includegraphics[width=\textwidth]{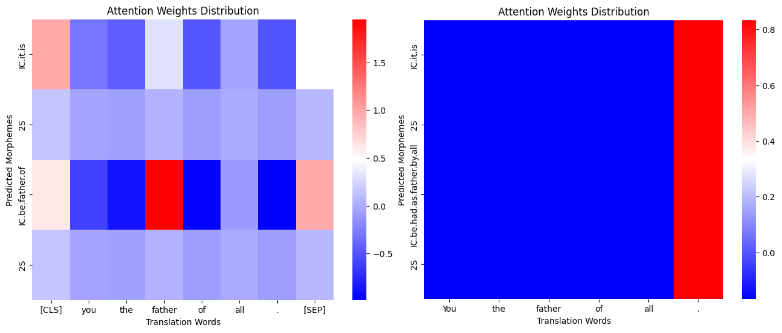}
    \caption{Difference from mean attention weights of glossed output tokens (y-axis) with respect to encoded translation tokens (x-axis) for an Arapaho example (attention weights are derived from the model BERT+attn+chr (left) and the model LSTM+attm (right)). The gold-standard glosses for this sentence: IC.it.is-2S IC.be.had.as.father.by.all-2S. }
    \label{fig:arpexample}
\end{figure*}

\begin{figure*}[htbp]
    \centering
    \includegraphics[width=\textwidth]{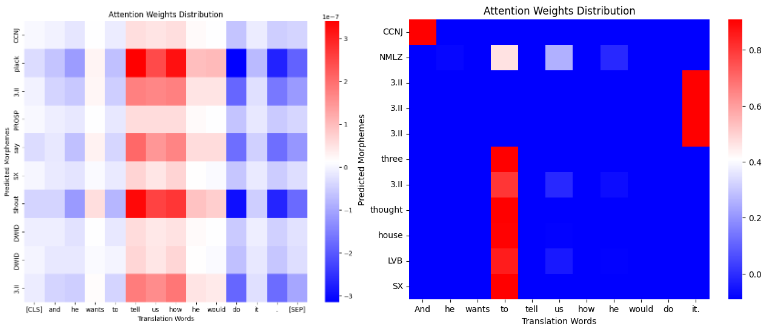}
    \caption{Difference from mean attention weights of glossed output tokens (y-axis) with respect to encoded translation tokens (x-axis) for a Gitksan example (attention weights are derived from the model BERT+attn+chr (left) and the model LSTM+attm (right)). The gold-standard glosses for this sentence: CCNJ want-3.II PROSP-3.I tell-T-3.II OBL-1PL.II MANR LVB-3.II.}
    \label{fig:gitexample}
\end{figure*}

\begin{figure*}[htbp]
    \centering
    \includegraphics[width=\textwidth]{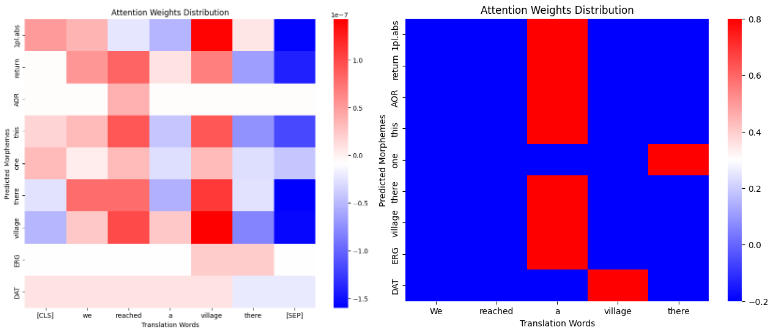}
    \caption{Difference from mean attention weights of glossed output tokens (y-axis) with respect to encoded translation tokens (x-axis) for a Lezgi example (attention weights are derived from the model BERT+attn+chr (left) and the model LSTM+attm (right)). The gold-standard glosses for this sentence: 1pl.abs return-AOR this one there village-ERG-DAT. }
    \label{fig:lezexample}
\end{figure*}

\begin{figure*}[htbp]
    \centering
    \includegraphics[width=\textwidth]{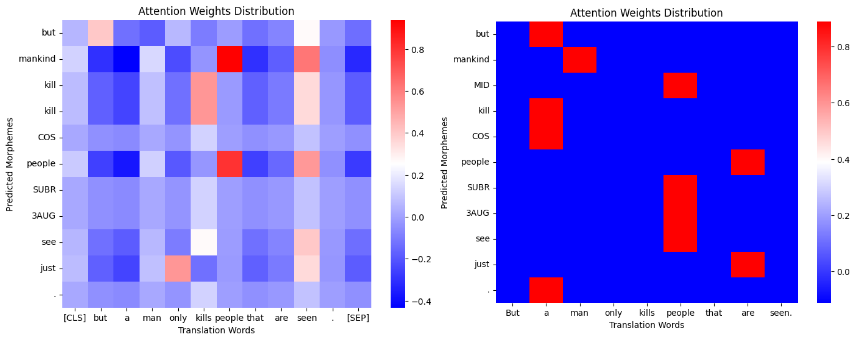}
    \caption{Difference from mean attention weights of glossed output tokens (y-axis) with respect to encoded translation tokens (x-axis) for a Nat\"ugu example (attention weights are derived from the model BERT+attn+chr (left) and the model LSTM+attm (right)). The gold-standard glosses for this sentence: but mankind MID-kill-COS-3MINIS people SUBR PAS-see-INTS-just.}
    \label{fig:ntuexample}
\end{figure*}

\begin{figure*}[htbp]
    \centering
    \includegraphics[width=\textwidth]{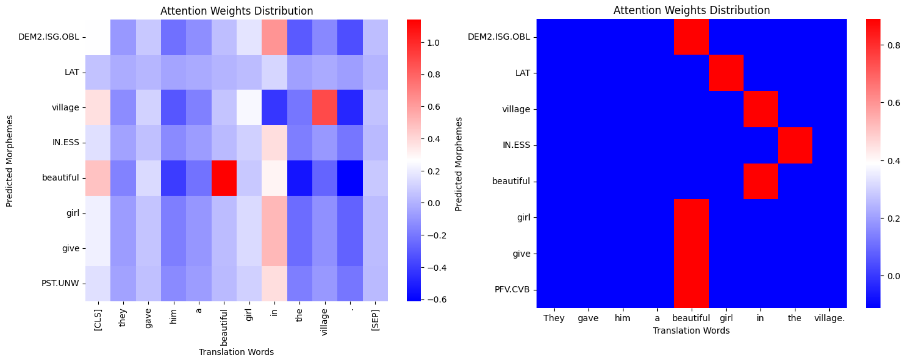}
    \caption{Difference from mean attention weights of glossed output tokens (y-axis) with respect to encoded translation tokens (x-axis) for a Tsez example (attention weights are derived from the model BERT+attn+chr (left) and the model LSTM+attm (right)). The gold-standard glosses for this sentence: DEM2.ISG.OBL-LAT village-IN.ESS beautiful girl give-PST.UNW}
    \label{fig:ddoexample}
\end{figure*}

\begin{figure*}[htbp]
    \centering
    \includegraphics[width=\textwidth]{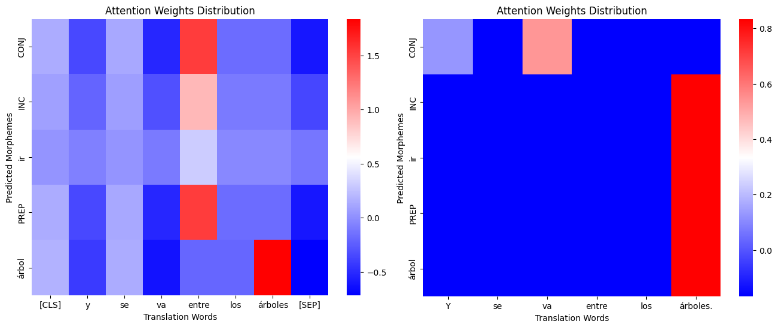}
    \caption{Difference from mean attention weights of glossed output tokens (y-axis) with respect to encoded translation tokens (x-axis) for a Uspanteko example (attention weights are derived from the model BERT+attn+chr (left) and the model LSTM+attm (right)). The gold-standard glosses for this sentence: CONJ INC-ir PREP árbol.}
    \label{fig:uspexample}
\end{figure*}

To assess whether our model is able to successfully incorporate translation information, we visualize attention patterns (from the BERT+attn+chr model) over the English translation representations. Figure \ref{fig:attdis-natugu-bert1} presents an example for Nat\"ugu. Attention weights are displayed in a heat map, where each cell indicates  difference from mean attention: $a - 1/(n+2)$. Here $n$ is the length of the translation in tokens ($+2$ here because of the start-of-sequence and end-of-sequence tokens \textsc{[cls]} and \textsc{[sep]} which 
are concatenated to the translation). 
 Positive red cells inidicate high attention and negative blue cells low attention. The visualization clearly indicates that the model attends to the relevant tokens in the translation when predicting the stems {\it people}, {\it mankind} and {\it kill}. Figure \ref{fig:arpexample}-Figure \ref{fig:uspexample} shows randomly picked heat maps for the rest of the languages. We can see that attention weights for the larger shared task datasets tend to express relevant associations, while attention weights for the ultra low-resource training sets largely represent noise. Figure \ref{fig:arpexample}-Figure \ref{fig:uspexample} also displays attention distributions when translations are incorporated using a randomly initialized LSTM instead of a pre-trained language model. These distributions also largely represent noise indicating that pre-trained models confer an advantage. 

\section{Prompt template}
\label{template}

\noindent You are a linguistic annotator for the Gitksan language, tasked with correcting errors in glossing based on translation details and morpheme translations. Your task is to adjust errors in the stems (in lowercase) without changing the total number of morphemes or words in the gloss. Each gloss element is separated by hyphens within morphemes and spaces between words.

\bigskip

\noindent Here are two examples:

\noindent \textbf{Example 1:} Gitksan sentence is \{example['train1-raw-sentence']\}. You are provided with morpheme translations according to the dictionary: \{example['train1-word/morpheme-translation']\}. The English translation for this sentence is: \{example['train1-sentence-translation']\}. The glossing pending to be revised is: \{example['train1-silver-gloss']\}. The corrected gloss is \{example['train1-gold-gloss']\}.

\bigskip

\noindent \textbf{Example 2:} Gitksan sentence is \{example['train2-raw-sentence']\}. You are provided with morpheme translations according to the dictionary: \{example['train2-word/morpheme-translation']\}. The English translation for this sentence is: \{example['train2-sentence-translation']\}. The glossing pending to be revised is: \{example['train2-silver-gloss']\}. The corrected gloss is \{example['train2-gold-gloss']\}.

\bigskip

\noindent Now, here's the gloss you need to correct:

\noindent Gitksan sentence is \{example['test-raw-sentence']\}. You are provided with morpheme translations according to the dictionary: \{example['test-word/morpheme-gloss']\}. The English translation for this sentence is: \{example['test-translation']\}. The glossing pending to be revised is: \{example['test-silver-gloss']\}. 

\noindent What is the corrected gloss for this sentence? You should answer in this format: \textbf{The corrected gloss is:} (your generated answer). Note, don't change the total number of words or morphemes in the gloss.

\section{Lexical Morpheme Accuracy}
\label{lex-technique}
Here we only evaluate the lexical morpheme accuracy. Results are shown in Table \ref{app:promptlex_stats1-lex}.

\begin{table}
\begin{adjustbox}{width=\columnwidth,center}
\begin{tabular}{lllllll}
\toprule
Model setting        & arp  & lez  & ntu  & ddo & usp & git      \\\midrule
T5/BERT+attn+chr & 83.68  & 81.29 & 81.51 & \textbf{92.79} & \textbf{82.75} & 12.83  \\
+GPT4-random & 84.78 &85.12 &83.19  & 90.52 & 70.54 &  26.79 \\
+GPT4-BERT-Sim & 85.13&\textbf{86.35} &83.33  & 91.23 & 73.28 & 27.13  \\
+GPT4-Overlap&\textbf{86.54} &86.20 & 84.17  & 91.76 & 74.91  & 27.17  \\
+GPT4-LCS & 85.97 &85.86 & \textbf{84.87} & 90.87 & 73.65 & 26.98 \\
+LLaMA3-Overlap &85.23 & 84.05 & 83.88 & 89.54 & 71.43  & \textbf{29.81}  \\\bottomrule
\end{tabular}
\end{adjustbox}
\caption{Lexical morpheme accuracy across languages in the 2023 Sigmorphon Shared Task \cite{ginn2023findings}  with `arp' representing Arapaho, `git' for Gitksan, 
 `lez' for Lezgi, `ntu' for Nat\"ugu, `ddo' for Tsez, and `usp' for Uspanteko. Model specifics are elaborated in Section \ref{experiment}. }
\label{app:promptlex_stats1-lex}
\end{table}


\section{BERT score}
\label{bert-score}
Specifically, we compare tokens using BERT embeddings and calculate similarity scores with the BERT model. The results are shown in Table \ref{prompt-bert}. As we do not have access to the results from \newcite{girrbach-2023-tu}, we use the LSTM-encoder classifier model as our baseline instead. The BERT score results align closely with the word-level accuracy.
\begin{table}[]
\begin{adjustbox}{width=0.5\textwidth,center}
\begin{tabular}{lllllll}
\toprule
Model setting        & arp  & lez  & ntu  & ddo & usp & git     \\\midrule
LSTM & 0.889  &0.873  & 0.826 & 0.925 & 0.783 & 0.434   \\
T5/BERT+attn+chr & 0.895  & 0.913 & 0.860   & \textbf{0.942} & \textbf{0.864}  & 0.468   \\
T5/BERT+attn+chr+Prmpt & \textbf{0.896} &\textbf{0.922} & \textbf{0.862} &  0.940 & 0.807 & \textbf{0.526}  \\\bottomrule
\end{tabular}
\end{adjustbox}
\caption{BERT score of lexical morphemes of languages in the 2023 Sigmorphon Shared Task \cite{ginn2023findings}, with `arp' representing Arapaho, `git' for Gitksan, 
 `lez' for Lezgi, `ntu' for Nat\"ugu, `ddo' for Tsez, and `usp' for Uspanteko. Model specifics are elaborated in Section \ref{experiment}. }
\label{prompt-bert}
\end{table}

\section{Dictionary Information}
\label{dictionary-source}
The Arapaho dictionary was accessed from \url{https://homewitharapaho.wordpress.com/wp-content/uploads/2015/03/arapaho-dictionary1.pdf}.

The Gitksan dictionary is downloaded from \url{http://www.gitxsansimalgyax.com/dictionaries.html}.

Lezgi data is unpublished and obtained through personal communication with a linguist.

Word number information of these dictionaries are in Table \ref{app:prompt-word-dict}.


\begin{table}

\begin{adjustbox}{width=\columnwidth,center}
\begin{tabular}{lll}
\toprule
Language        & total words(num) & new words(num)    \\\midrule
Arapaho& 2436  &  2155    \\
Lezgi&2081 &1299  \\
Gitksan&2034 &2019\\\bottomrule
\end{tabular}
\end{adjustbox}
\caption{The table details the dictionary information for Arapaho, Lezgi, and Gitksan, including the number of total words and the number of new words compared with the training data.}
\label{app:prompt-word-dict}
\end{table}

\end{document}